%% file: main.tex
\documentclass[sigconf, screen]{acmart}

\usepackage{graphicx}
\usepackage{booktabs}
\usepackage{array}
\usepackage{makecell}
\usepackage{xcolor}

\usepackage{dingbat}
\usepackage{multirow}
\usepackage{tabularx}
\usepackage{colortbl}
\usepackage{pifont}
\AtBeginDocument{%
  }

\setcopyright{acmlicensed}
\copyrightyear{2025}
\acmYear{2025}
\acmDOI{XXXXXXX.XXXXXXX}
\acmConference[Conference acronym 'CIKM]{The Conference on Information and Knowledge Management}{Nov 10--14, 2025}{Seoul, KR}
\acmISBN{978-1-4503-XXXX-X/2018/06}

\begin{document}

\title{Can Large Vision-Language Models Understand\\ Multimodal Sarcasm?}


\author{Xinyu Wang}
\authornote{Both authors contributed equally to this research.}
\affiliation{
    \institution{The University of Texas at Dallas}
    \city{Richardson}
    \country{USA}
}
\email{cpwxyxwcp@gmail.com}

\author{Yue Zhang}
\authornotemark[1]
\affiliation{
    \institution{The University of Texas at Dallas}
    \city{Richardson}
    \country{USA}
}
\email{yue.zhang@utdallas.edu}

\author{Liqiang Jing}
\authornote{Liqiang Jing is the corresponding author.}
\affiliation{
    \institution{The University of Texas at Dallas}
    \city{Richardson}
    \country{USA}
}
\email{jingliqiang6@gmail.com}

\renewcommand{\shortauthors}{Xinyu Wang, Yue Zhang, Liqiang Jing}


\input{Section/0_Abstract}



\received{20 February 2007}
\received[revised]{12 March 2009}
\received[accepted]{5 June 2009}

\maketitle
\input{Section/1_Introduction}
\input{Section/2_Related-Works}
\input{Section/3_Test-Task}

\input{Section/4_Experimental-Result}

\input{Section/5_Method}

\input{Section/6_Conclusion}


\bibliographystyle{ACM-Reference-Format}
\bibliography{main}



\end{document}

%% file: Section/0_Abstract.tex
\begin{abstract}

Sarcasm is a complex linguistic phenomenon that involves a disparity between literal and intended meanings, making it challenging for sentiment analysis and other emotion-sensitive tasks. While traditional sarcasm detection methods primarily focus on text, recent approaches have incorporated multimodal information. However, the application of Large Visual Language Models (LVLMs) in Multimodal Sarcasm Analysis (MSA) remains underexplored. In this paper, we evaluate LVLMs in MSA tasks, specifically focusing on Multimodal Sarcasm Detection and Multimodal Sarcasm Explanation. Through comprehensive experiments, we identify key limitations, such as insufficient visual understanding and a lack of conceptual knowledge. To address these issues, we propose a training-free framework that integrates in-depth object extraction and external conceptual knowledge to improve the model's ability to interpret and explain sarcasm in multimodal contexts. The experimental results on multiple models show the effectiveness of our proposed framework. The code is available at \url{https://github.com/cp-cp/LVLM-MSA}.

\end{abstract}

%% file: Section/1_Introduction.tex
\section{Introduction}
Sarcasm is a linguistic phenomenon where the intended meaning opposes the literal interpretation, commonly used to convey sharp criticism or mockery toward someone or something. Understanding sarcasm is important for tasks like sentiment analysis, social media analysis, and customer feedback service, as these tasks require accurately identifying and interpreting people's emotions. Traditional sarcasm detection methods \cite{joshi-etal-2016-word, amir-etal-2016-modelling} primarily focus on text modality. With the development of multimedia, recent approaches \cite{Pattern, MSDD, Incongruity} shift research attention into utilizing multimodal information to conduct Multimodal Sarcasm Analysis (MSA).

Despite the existing advancements in MSA \cite{MSD, MORE}, the performance of LVLMs in MSA remains unexplored. 
Recently, LVLMs have been evaluated on various tasks, such as VQA \cite{MME}, visual entailment \cite{DBLP:conf/aaai/ZhangJG25, DBLP:journals/corr/abs-2506-22385}, sentiment analysis \cite{sentiment1,sentiment2,sentiment3,sentiment4}, and multimodal summarization \cite{DBLP:journals/ijautcomp/JingLXYSS23,DBLP:conf/sigir/LinJSLSN23,DBLP:conf/sigir/SongJLZCN22,zhang2024fine}, demonstrating impressive capabilities in the understanding of both visual and textual modalities~\cite{MME, LVLM-eHub, Qwen}.  
MSA is a non-trivial task because it involves understanding subtle cultural, emotional, and contextual nuances that are not always explicitly stated. Additionally, the task requires not only a deep understanding of both textual and visual information but also the ability to effectively leverage and integrate these modalities, further increasing the complexity of the problem. Due to the above reason, exploring the performance of LVLMs on MSA is vital for comprehensively evaluating their abilities on multimodal understanding.

\begin{figure}[!t]
    \centering
    \includegraphics[width=\linewidth]{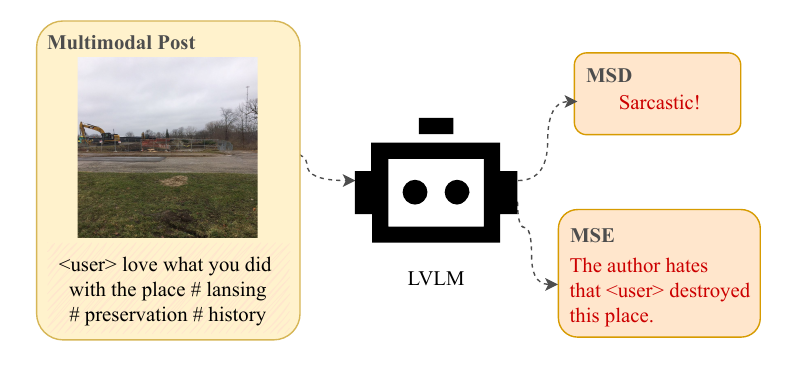}
    \label{fig:intro}
    \vspace{-3mm}
    \caption{Multimodal Sarcasm Analysis, including MSD and MSE.}
    \vspace{-3mm}
\end{figure}




Therefore, our first research question is \textit{\textbf{RQ1: what is the zero-shot performance on MSA for LVLMs?}}
To explore this research question, we assess LVLMs' performance on two key MSA tasks: Multimodal Sarcasm Detection (MSD) and Multimodal Sarcasm Explanation (MSE). 
Through comprehensive experiments, we found that LVLMs show poor zero-shot performance on MSA tasks. 
Therefore, we then explore the second research question \textit{\textbf{RQ2: how to improve the performance on MSA for LVLMs without fine-tuning?}}
Unlike other training-based methods\cite{prompt-tuning}, we focus on finding a train-free method to improve the effect.
We found that poor ability stems from limited visual understanding and a lack of conceptual knowledge. 
To address these issues, we propose an effective framework that enhances model performance by integrating in-depth object extraction and external conceptual knowledge, thereby improving sarcasm interpretation and explanation in multimodal contexts. 

Our contributions can be summarized as: 
\begin{itemize}
    \item We categorize MSA into classification and explanation tasks, with the corresponding subtasks being MSD and MSE, and evaluate the capabilities of LVLMs on these two tasks to showcase their zero-shot cross-modal analysis abilities.
    \item We revisit previous studies on LVLMs and analyze the limitations of LVLMs in handling sarcasm, identifying two key challenges: limited visual capabilities and insufficient conceptual knowledge.
    \item We propose a multi-source semantic-enhanced multimodal sarcasm understanding framework that improves LVLMs' sarcasm understanding by incorporating external knowledge sources, including detailed object and conceptual knowledge. The experimental results across multiple LVLMs show the effectiveness of our framework. Furthermore, this method provides a new perspective for extending LVLMs in complex multimodal tasks.
\end{itemize}


    
    

%% file: Section/2_Related-Works.tex
\section{Related Work}

\subsection{Evaluation in LVLMs.}
LVLMs emerged as a focal point of interest due to their remarkable
capabilities in handling diverse multimodal tasks. Researchers have developed various LVLMs \cite{LLaVA, MiniGPT4, InstructBLIP}, which generally consist of an image encoder and a text decoder derived from pre-trained models and a text-image alignment module. 
These LVLMs exhibit excellent generalization abilities and can be applied to many scenarios. 
As multimodal research deepens, some studies began focusing on evaluating the zero-shot capabilities of LVLMs\cite{lu2023evaluation}, forming a series of benchmarks\cite{MM-Instruct, Goat-Bench}. In addition, more and more studies have found that the existing LVLMs have some defects, such as Hallucinations\cite{FaithScore}, and Overfitting \cite{LVLM-eHub}. To mitigate these issues, techniques like Chain-of-Thought \cite{MM-CoT, DDCOT} and In-Context-Learning \cite{VICL, EvalICL} are increasingly employed to enhance model performance during inference. Despite extensive research into the evaluation and enhancement of LVLMs, there is a limited focus on assessing and improving their performance for sarcasm analysis tasks.

\subsection{Multimodal Sarcasm Detection and Explanation.}
In the sarcasm analysis task, there are two key sub-tasks, including sarcasm detection and sarcasm explanation. Recent studies \cite{MORE, MSDD, DBLP:conf/naacl/TangLY024, Cofipara, Wen_2023_CVPR, Incongruity, KnowleNet, ModelIncongruity, TEAM, DBLP:journals/corr/abs-2504-21226, tong2025rainbow,DBLP:journals/corr/abs-2402-03658} have concentrated on developing specialized training methods and models to achieve state-of-the-art results in these tasks. Notably, some research efforts, such as those presented in \cite{TEAM, Wisdom}, have demonstrated that incorporating additional external knowledge—such as contextual information or background knowledge—can significantly enhance model performance by providing a richer understanding of the sentiment conveyed.
Different from the existing works, we focus on evaluating and improving sarcasm understanding ability in zero-shot scenarios.

%% file: Section/3_Test-Task.tex
\section{Test Task}

To methodically assess the MSA abilities of LVLMs, we conduct a comprehensive evaluation focusing on their understanding and grounding capabilities in sarcasm detection and explanation. Specifically, this study addresses two tasks: Multimodal Sarcasm Detection (MSD)~\cite{SarcasmDetection}, and Multimodal Sarcasm Explanation (MSE)~\cite{MORE}.

\subsection{Multimodal Sarcasm Detection}

\paragraph{Task Formulation}
Suppose that we have a set of $N$ testing samples $\mathcal{D}=\{s^1,s^2,\cdots,s^N\}$. Each samples $s^i = (\mathcal{T}^i,I^i,Y^i)$ involves three elements. Here, $\mathcal{T}^i$ denotes the textual sentence, $I^i$ denotes the image, and $Y^i$ is the ground truth label for the $i$-th sample.
The MSD task aims to test whether a model ${\mathcal{F}}$ can precisely identify sarcasm in  a given text and its attached image as follows,
\begin{equation}  
	\hat{Y}^i={\mathcal{F}}(\mathcal{T}^i,\ I^i) ,
\end{equation}
where $\hat{Y}^i$ is the binary classification prediction result of ${\mathcal{F}}$. 

\paragraph{Metrics}
To evaluate the performance of LVLMs in the MSD task, we utilize Accuracy and F1-Score as the evaluation metrics.


\subsection{Multimodal Sarcasm Explanation}

Suppose we have a testing dataset $\mathcal{D}$ composed of $N$ samples, $\mathcal{D}=\{d_1,d_2,\cdots,d_N\}$. Each sample $d_i=\{T_i,I_i,Y_i\}$, where $T_i$ denotes the input sentence, $I_i$ is the input image, and $Y_i$ denotes the target explanation text. The target of this task is to test whether a model $\mathcal{F}$ is able to generate the sarcasm explanation based on the given multimodal input as follows,
\begin{equation}
    \hat{Y_i} = \mathcal{F}(T_i,I_i),
\end{equation}
where $\hat{Y_i}$ is the generated explanation text by 
 $\mathcal{F}$. 
 



\paragraph{Metrics}
For evaluating the performance of LVLMs in the Multimodal Sarcasm Explanation (MSE) task, we utilize {BLEU} \cite{BLEU}, ROUGE\cite{ROUGE}, and METEOR \cite{METEOR},  which are typically used in explanation tasks.






\begin{table}[]
    \caption{Evaluation of LVLMs on the MSD task. The table shows accuracy and F1 scores for each model, with the best results highlighted in bold.}
    \label{RTR}
    \centering
    \begin{tabular}{l|ll}
        \toprule
        \textbf{Model} & \textbf{Accuracy (\%)} & \textbf{F1 (\%)} \\ 
        \midrule
        LLaVA         & 41.1  & 57.4  \\ 
        MiniGPT      & 44.2  & 54.5  \\ 
        InstructBLIP  & 42.5 & 55.2  \\ 
        GPT-4o & \textbf{65.9} & \textbf{68.6} \\
        \bottomrule
    \end{tabular}
\end{table}

\begin{table}[]
\caption{Evaluation of LVLMs on the MSE task. The table presents BLEU (B1, B2, B3, B4), ROUGE (RL, R1, R2), and Mentor scores for each model, with the highest scores in each category highlighted in bold}
\label{ETR}
\centering
\resizebox{\columnwidth}{!}{
\begin{tabular}{l|llll|lll|l}
\toprule
\multirow{2}*{\textbf{Model}}   & \multicolumn{4}{c}{\textbf{BLEU}}    & \multicolumn{3}{|c|}{\textbf{Rouge}}   & \multirow{2}*{\textbf{Mentor}} \\
\cmidrule{2-8}
& B1 & B2 & B3 & B4 & RL  & R1 & R2 &  \\ 
\midrule
LLaVA & 8.285  & 4.689  & 3.120  & 2.203  & \textbf{14.500}  & 14.488  & 5.068  & \textbf{26.628} \\ 
InstructBLIP & \textbf{9.617}  & \textbf{7.265}  & \textbf{5.823}  & \textbf{4.773}  & 7.738   & 9.133   & \textbf{5.183}  & 7.901 \\      
MiniGPT & 7.252 & 2.928 & 1.587 & 1.007 & 12.295 & \textbf{16.441} & 2.403 & 7.776\\
GPT-4o            & 5.803 & 3.423 & 2.315 & 1.638 & 10.778 & 12.143 & 4.463 & 25.115 \\
\bottomrule
\end{tabular}
\vspace{-5mm}
}
\end{table}

%% file: Section/4_Experimental-Result.tex
\section{What is Performance of LVLMs on Sarcasm Tasks?}
\makeatletter
\patchcmd{\@makecaption}
  {\scshape}
  {}
  {}
  {}
\makeatother

\subsection{Datasets}

For the MSD and MSE tasks, we selected the testing sets of  MSDD~\cite{MSDD} and MORE~\cite{MORE} as evaluation sets, respectively. MSDD is comprised of multimodal posts from Twitter, where each post incorporates textual content and an accompanying image. Each post is assigned a label from the predefined set \{sarcastic, unsarcastic\}. MORE dataset collected sarcastic posts from existing multimodal sarcasm detection datasets. The researchers carefully checked the collected posts and annotated the explanation for each post.
Statistics of our testing sets are summarized in Table \ref{status}.

\begin{table}[]
    \caption{Statistics of our evaluation sets.}
    \label{status}
    \centering
    \begin{tabular}{llll}
        \toprule
        \textbf{Task}  & \textbf{Sources}  & \textbf{Distribution} & \textbf{Total} \\ 
        \midrule
        \multirow{2}{*}{MSD}  & \multirow{2}{*}{MSDD} & Sarcastic 959& \multirow{2}{*}{2409}\\
        & & Unsarcastic 1450 & \\ 
        \cline{1-4}
        \multirow{2}{*}{MSE}& \multirow{2}{*}{MORE}& Caption Avg.length = 19.43& \multirow{2}{*}{352} \\ 
        & &Explanation Avg.length =  15.08 & \\
        \bottomrule
    \end{tabular}
    \vspace{-5mm}
\end{table}

\subsection{Models}

To gain a comprehensive understanding of the current state of LVLMs in sarcasm analysis, we select 3 open-source LVLMs and 1 closed-source LVLM: 
(1) \textbf{LLaVA-v1.5} \cite{LLaVA} is an enhanced version of LLaVA integrates the visual encoder CLIP with the language model LLaMA, optimized for comprehensive visual and linguistic understanding and finally refined through instruction tuning using image-based linguistic data generated by GPT-4 \cite{GPT4}. We select llava-v1.5-7b \footnote{\url{https://github.com/haotian-liu/LLaVA}.} for evaluation.
(2) 
\textbf{MiniGPT} \cite{MiniGPT4} is an open-source LVLM that aligns a frozen visual encoder with a frozen language model LLaMA \cite{LLaMA}, refined through instruction tuning using some instruction datasets. We utilize minigptv2-llama-7b \footnote{\url{https://github.com/Vision-CAIR/MiniGPT-4}.} for evaluation.
(3)
\textbf{InstructBLIP} \cite{InstructBLIP} is 
an open-source LVLM based on a pre-trained BLIP-2 model, achieving multimodal capabilities through visual-language instruction adjustment. We utilize the InstructBLIP-vicuna-7b \footnote{\url{https://github.com/salesforce/LAVIS/tree/main/projects/instructblip}.} for testing.
(4)
\textbf{GPT-4o} \cite{GPT4} is a version of the GPT-4 model designed for optimized performance in both language and vision tasks. It has been refined through stages of pre-training, instruction tuning, and reinforcement learning from human feedback. We utilize gpt-4o-mini version for evaluation.



\begin{figure*}[!h]
    \centering
    \includegraphics[width=0.7\linewidth]{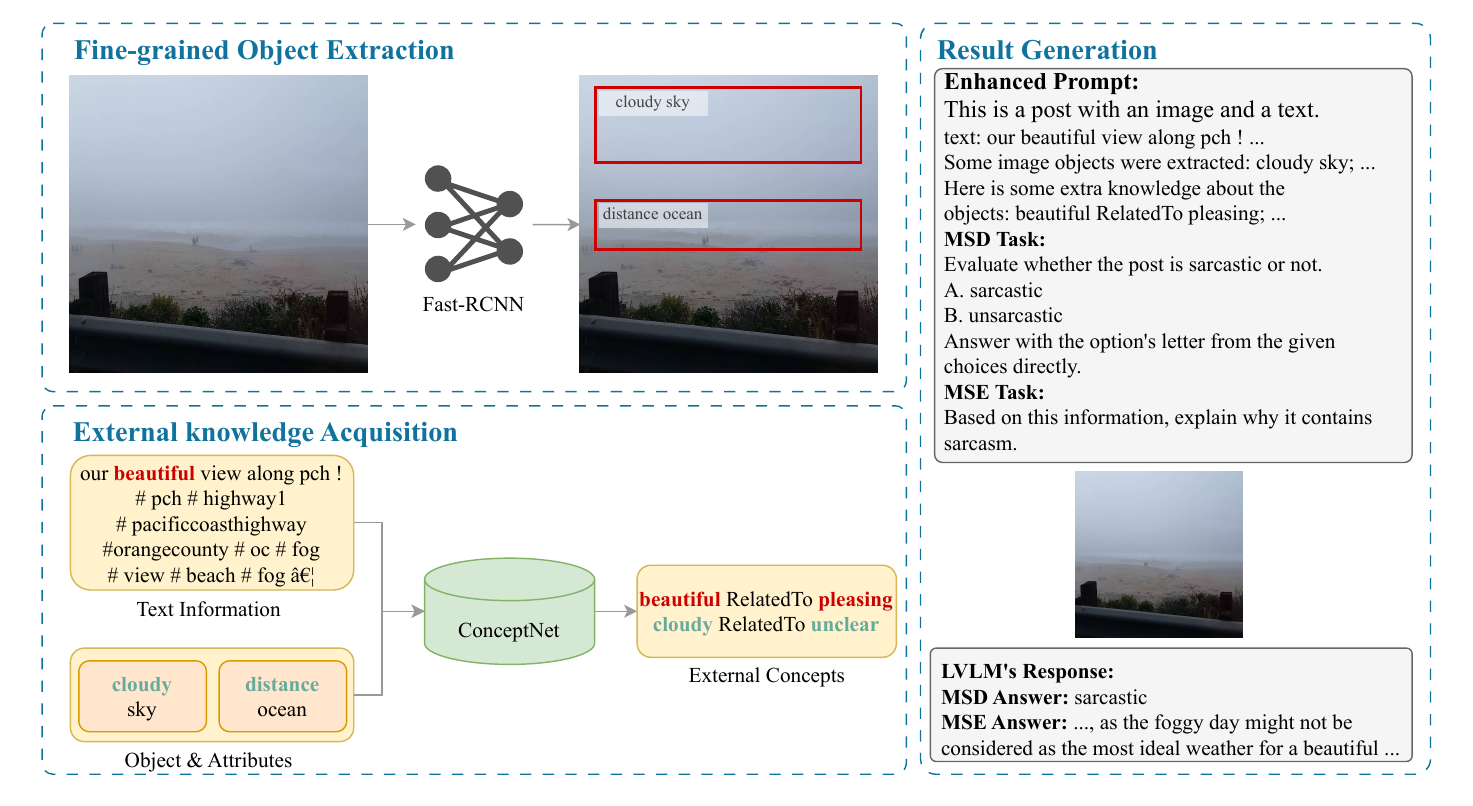}
    \vspace{-3mm}
    \caption{\textbf{The multi-source semantic enhanced sarcasm understanding framework}: 1) \textbf{Fine-grained object extraction} using Fast-RCNN to detect objects from images, 2) \textbf{External knowledge acquisition}, linking text features and image objects to external concepts through ConceptNet, enriching understanding by associating attributes. 3) \textbf{Result generation} for sarcasm detection and explanation. The framework leverages these components to enhance sarcasm comprehension in multimodal contexts.}
    \label{method}
    \label{fig:method}
    \vspace{-4mm}
\end{figure*}

\subsection{Result}
In Table \ref{RTR} and \ref{ETR}, we showcase the performance of all 4 LVLMs tested in the zero-shot setting on MSE and MSD tasks. Based on the results in Table \ref{RTR} and \ref{ETR}, we draw the following observations:
On the whole, GPT-4o shows the strongest performance on the MSD task, achieving the highest accuracy and F1 score, while it demonstrated poor performance in sarcasm explanation generation. 
Other models perform poorly on the sarcasm detection task and show different capabilities on the sarcasm explanation task, highlighting the need for further improvements to help them achieve better results.

%% file: Section/5_Method.tex
\section{Multi-source Semantic enhanced Multimodal Sarcasm Understanding}


\subsection{Overview}


To enhance performance on multimodal sarcasm detection and explanation tasks, we revisited previous studies on LVLMs and identified two primary challenges that may have contributed to the poor performance of these LVLMs:
     1) \textbf{Lack of Vision Ability}:
    LVLMs may produce hallucinations \cite{Hallucination, FaithScore, lu2023evaluation, AnalyzeHallucination}, especially in fine-grained objects, which leads to their limited capabilities in visual understanding.
    2) \textbf{Lack of Potential External Knowledge}:
    Due to the limited nature of the training data, LVLMs may not be able to make connections between existing visual entities and related concepts \cite{FakeNews, LEMMA, Wisdom}, such as sentiment knowledge, which is important for sarcasm understanding.

To address both challenges, we propose a multi-source semantic enhanced multimodal sarcasm understanding framework, which mainly consists of three steps: fine-grained object extraction, external knowledge acquisition, and result generation. 
Fig. \ref{method} illustrates the overview of our method.
In this framework, we focus on enhancing the performance of LVLMs in MSD and MSE tasks by introducing recognized objects with attributes and external knowledge. This approach aims at improving the models' ability to understand and interpret the potential meaning conveyed in images and text, especially in cases where the context is complex or subtle. 
\subsection{Method}

\begin{table*}[t]
    \centering
\resizebox{1.4\columnwidth}{!}{
    \begin{tabular}{l|l|llll|lll|l}
    \toprule
        \multirow{2}*{\textbf{Model}} & \multirow{2}*{\textbf{Method}} &\multicolumn{4}{c}{\textbf{BLEU}} & \multicolumn{3}{|c|}{\textbf{Rouge}} & \multirow{2}*{\textbf{Mentor}} \\
        \cmidrule{3-9}
        ~ &~ & B1 & B2 & B3 & B4 & RL & R1 & R2 & ~ \\ \midrule
        
        \multirow{2}*{LLaVA}
        &Baseline  & 8.285  & 4.689  & 3.120  & 2.203  & 14.500  & 14.488  & 5.068  & 26.628 \\ 
        &+Ours    & \textbf{9.430}$^*$ & \textbf{5.681}$^*$ & \textbf{3.925}$^*$ &  \textbf{2.852}$^*$ & \textbf{16.301}$^*$ & \textbf{16.549}$^*$ & \textbf{6.447}$^*$ & \textbf{29.640}$^*$\\  
        \midrule
        
        \multirow{2}{*}{InstructBLIP}
        &Baseline         & 9.617  & 7.265  & 5.823  & 4.773  & 7.738   & 9.133   & 5.183  & 7.901 \\
        &+Ours & \textbf{14.196}$^*$ & \textbf{9.198}$^*$  & \textbf{6.866}$^*$  & \textbf{5.343}$^*$  & \textbf{22.473}$^*$  &\textbf{23.386}$^*$  & \textbf{11.140}$^*$  & \textbf{30.380}$^*$ \\
        \midrule
        
        \multirow{2}{*}{MiniGPT}
        &Baseline   & 7.252 & 2.928 & 1.587 & 1.007 & 12.295 & 16.441 & 2.403 & 7.776\\
        &+Ours   & \textbf{21.173}$^*$ & \textbf{11.400}$^*$ & \textbf{7.276}$^*$ & \textbf{4.846}$^*$ & \textbf{19.000}$^*$ & \textbf{23.074}$^*$ & \textbf{6.202}$^*$ & \textbf{16.600}$^*$\\
        \midrule
        
        \multirow{2}{*}{GPT-4o}
        &Baseline   & 5.803 & 3.423 & 2.315 & 1.638 & 10.778 & 12.143 & 4.463 & 25.115 \\
        &+Ours   & \textbf{8.863}$^*$ & \textbf{4.989}$^*$ & \textbf{3.202}$^*$ & \textbf{2.148}$^*$ & \textbf{14.242}$^*$ & \textbf{15.475}$^*$ & \textbf{5.595}$^*$ & \textbf{28.717}$^*$\\
        \bottomrule
    \end{tabular}
}
    \caption{Performance comparison of LVLMs on the MSE task. The best results are highlighted in bold. * indicates that the p-value of the significance test comparing our result with the best baseline result is less than 0.01.}   \label{MSE}
    \vspace{-2mm}
\end{table*}

\paragraph{Fine-grained Object Extraction}
We extract objects and their attributes from the image and incorporate them into the prompt, providing LVLMs with a more comprehensive context to leverage for sarcasm understanding.
To extract fine-grained visual elements from images, we employ Fast-RCNN \cite{FastRCNN} as one feasible implementation for object recognition. Fast-RCNN has proven effective in identifying and describing objects with key visual attributes such as shape and color, which are essential for constructing a detailed representation of the image. Our intention is not to position Fast-RCNN as an alternative to more advanced models like CLIP or BERT; rather, it functions as a supporting component to enrich the input prompt with explicit object-level details. These enriched prompts can then help LVLMs better understand the implicit or contextual meaning conveyed by the image. We emphasize that Fast-RCNN is merely one of many possible tools for this role, chosen here for its practical advantages in our specific pipeline.
\paragraph{External Knowledge Acquisition}
While extracting objects and their attributes is essential for understanding the visual content, it alone may not suffice for accurate sarcasm analysis. This is because LVLMs often struggle to interpret the deeper semantics or contextual implications of the visual elements and accompanying text.
To address this, we incorporate external conceptual knowledge that helps connect the visual attributes to higher-level meanings. Rather than relying on unstructured knowledge retrieval techniques such as RAG—whose primary design is for open-domain text generation and question answering—we opt for a structured commonsense knowledge base, namely ConceptNet \cite{ConceptNet}.
ConceptNet offers explicit, semantically rich relationships between concepts, enabling us to map both object attributes and textual elements to their neighboring concepts, such as “hospital” → “sickness” or “red” → “warning”. These connections are especially beneficial in sarcasm detection, where indirect implications often hinge on such latent associations.
In contrast to RAG, which may retrieve loosely related or verbose textual passages, ConceptNet enables more precise and interpretable enrichment of the input. We emphasize that our goal is not general knowledge augmentation, but grounded conceptual linking between visual cues and abstract notions, for which structured knowledge graphs are more effective.
\paragraph{Result Generation}
After acquiring these concepts, we incorporate them into the prompt to provide the LVLMs with a more nuanced understanding of the objects and input text. This enables the models to interpret the sarcastic meaning expressed in both the image and the text more accurately. We use the enhanced prompt for the final sarcasm analysis, incorporating detailed object information and relevant external concepts.
\begin{table}[!ht]
    \vspace{-4mm}
    \centering
    \resizebox{0.65\columnwidth}{!}{
    \begin{tabular}{l|l|l|l}
        \toprule
        \textbf{Model}  & \textbf{Method} & \textbf{Accuracy (\%)} & \textbf{F1 (\%)} \\ 
        \midrule
        \multirow{2}{*}{LLaVA} 
        & Baseline             & 41.1  & 57.4 \\ 
        & +Ours         & \textbf{60.3} (+46.8\%)  & \textbf{59.4} (+3.5\%) \\
        \midrule
        \multirow{2}{*}{MiniGPT} 
        & Baseline     & 44.2 & 54.5 \\ 
        & +Ours     & \textbf{50.0} (+13.2\%)  & \textbf{56.5} (+3.7\%) \\ 
        \midrule
        \multirow{2}{*}{InstructBLIP} 
        & Baseline          & 42.5  & 55.2 \\ 
        & +Ours  & \textbf{51.5} (+21.2\%)  & \textbf{57.9} (+4.9\%) \\
        \midrule
        \multirow{2}{*}{GPT-4o} 
        & Baseline                 & 65.9 & 68.6 \\
        & +Ours         & \textbf{75.3} (+14.2\%)  & \textbf{72.1} (+5.1\%) \\
        \bottomrule
    \end{tabular}
    }
    \caption{Performance comparison of LVLMs on the MSD task. The table shows accuracy and F1 scores for each model using the baseline method and our method.}
    \label{MSD}
    \vspace{-12mm}
\end{table}
\subsection{Result}
We compare our method with existing baselines on each task and report the results in Table \ref{MSD} and Table \ref{MSE}. After introducing fine-grained objects and additional conceptual knowledge, our method has achieved the best results on both tasks. Impressively, in the MSD task, on GPT-4o, we achieved 75.3\% accuracy, a 14.2\% relative improvement over the baseline. In particular, on the MSE task, our method achieved great improvement on InstructBLIP and MiniGPT. In addition, we found that LLaVA, which had a poor baseline performance, was improved on both tasks after applying our method, which shows that our method effectively supplements visual capabilities and provides sufficient extra knowledge for the LVLMs to assist them in completing the sarcasm analysis task.
\vspace{-4mm}

%% file: Section/6_Conclusion.tex
\section{Conclusion}
In this paper, we evaluated the zero-shot capabilities of Large Vision-Language Models (LVLMs) on the core tasks of Multimodal Sarcasm Analysis (MSA), specifically focusing on sarcasm detection and explanation. Our findings reveal that LVLMs perform poorly in understanding multimodal sarcasm content, particularly due to limitations in visual semantics and conceptual knowledge. To address these gaps, we proposed incorporating in-depth object information and external conceptual knowledge sources to enhance the models' performance. 
This approach offers a promising direction for improving the ability of LVLMs to handle complex sarcastic content in multimodal scenarios.

\section{Acknowledgments}
We want to thank our anonymous reviewers for their feedback. This work was supported by the OpenAI Researcher Access Program 0000006384.

%% file: main.bbl

\begin{thebibliography}{54}


\ifx \showCODEN    \undefined \def \showCODEN     #1{\unskip}     \fi
\ifx \showISBNx    \undefined \def \showISBNx     #1{\unskip}     \fi
\ifx \showISBNxiii \undefined \def \showISBNxiii  #1{\unskip}     \fi
\ifx \showISSN     \undefined \def \showISSN      #1{\unskip}     \fi
\ifx \showLCCN     \undefined \def \showLCCN      #1{\unskip}     \fi
\ifx \shownote     \undefined \def \shownote      #1{#1}          \fi
\ifx \showarticletitle \undefined \def \showarticletitle #1{#1}   \fi
\ifx \showURL      \undefined \def \showURL       {\relax}        \fi
\providecommand\bibfield[2]{#2}
\providecommand\bibinfo[2]{#2}
\providecommand\natexlab[1]{#1}
\providecommand\showeprint[2][]{arXiv:#2}

\bibitem[Amir et~al\mbox{.}(2016)]%
        {amir-etal-2016-modelling}
\bibfield{author}{\bibinfo{person}{Silvio Amir}, \bibinfo{person}{Byron~C. Wallace}, \bibinfo{person}{Hao Lyu}, {and} \bibinfo{person}{Paula Carvalho Mário~J. Silva}.} \bibinfo{year}{2016}\natexlab{}.
\newblock \bibinfo{title}{Modelling Context with User Embeddings for Sarcasm Detection in Social Media}.
\newblock
\showeprint[arxiv]{1607.00976}~[cs.CL]


\bibitem[Bai et~al\mbox{.}(2023)]%
        {Qwen}
\bibfield{author}{\bibinfo{person}{Jinze Bai}, \bibinfo{person}{Shuai Bai}, \bibinfo{person}{Shusheng Yang}, \bibinfo{person}{Shijie Wang}, \bibinfo{person}{Sinan Tan}, \bibinfo{person}{Peng Wang}, \bibinfo{person}{Junyang Lin}, \bibinfo{person}{Chang Zhou}, {and} \bibinfo{person}{Jingren Zhou}.} \bibinfo{year}{2023}\natexlab{}.
\newblock \showarticletitle{Qwen-VL: {A} Frontier Large Vision-Language Model with Versatile Abilities}.
\newblock \bibinfo{journal}{\emph{CoRR}}  \bibinfo{volume}{abs/2308.12966} (\bibinfo{year}{2023}).
\newblock
\showeprint[arXiv]{2308.12966}


\bibitem[Banerjee and Lavie(2005)]%
        {METEOR}
\bibfield{author}{\bibinfo{person}{Satanjeev Banerjee} {and} \bibinfo{person}{Alon Lavie}.} \bibinfo{year}{2005}\natexlab{}.
\newblock \showarticletitle{{METEOR:} An Automatic Metric for {MT} Evaluation with Improved Correlation with Human Judgments}. In \bibinfo{booktitle}{\emph{Proceedings of the Workshop on Intrinsic and Extrinsic Evaluation Measures for Machine Translation and/or Summarization@ACL}}. \bibinfo{publisher}{Association for Computational Linguistics}, \bibinfo{pages}{65--72}.
\newblock


\bibitem[Bouazizi and Ohtsuki(2016a)]%
        {Pattern}
\bibfield{author}{\bibinfo{person}{Mondher Bouazizi} {and} \bibinfo{person}{Tomoaki Ohtsuki}.} \bibinfo{year}{2016}\natexlab{a}.
\newblock \showarticletitle{A Pattern-Based Approach for Sarcasm Detection on Twitter}.
\newblock \bibinfo{journal}{\emph{{IEEE} Access}}  \bibinfo{volume}{4} (\bibinfo{year}{2016}), \bibinfo{pages}{5477--5488}.
\newblock


\bibitem[Bouazizi and Ohtsuki(2016b)]%
        {SarcasmDetection}
\bibfield{author}{\bibinfo{person}{Mondher Bouazizi} {and} \bibinfo{person}{Tomoaki Ohtsuki}.} \bibinfo{year}{2016}\natexlab{b}.
\newblock \showarticletitle{A Pattern-Based Approach for Sarcasm Detection on Twitter}.
\newblock \bibinfo{journal}{\emph{{IEEE} Access}}  \bibinfo{volume}{4} (\bibinfo{year}{2016}), \bibinfo{pages}{5477--5488}.
\newblock


\bibitem[Cai et~al\mbox{.}(2019)]%
        {MSDD}
\bibfield{author}{\bibinfo{person}{Yitao Cai}, \bibinfo{person}{Huiyu Cai}, {and} \bibinfo{person}{Xiaojun Wan}.} \bibinfo{year}{2019}\natexlab{}.
\newblock \showarticletitle{Multi-Modal Sarcasm Detection in Twitter with Hierarchical Fusion Model}. In \bibinfo{booktitle}{\emph{{ACL}}}. \bibinfo{publisher}{Association for Computational Linguistics}, \bibinfo{pages}{2506--2515}.
\newblock


\bibitem[Chen et~al\mbox{.}(2024)]%
        {Cofipara}
\bibfield{author}{\bibinfo{person}{Zixin Chen}, \bibinfo{person}{Hongzhan Lin}, \bibinfo{person}{Ziyang Luo}, \bibinfo{person}{Mingfei Cheng}, \bibinfo{person}{Jing Ma}, {and} \bibinfo{person}{Guang Chen}.} \bibinfo{year}{2024}\natexlab{}.
\newblock \showarticletitle{CofiPara: {A} Coarse-to-fine Paradigm for Multimodal Sarcasm Target Identification with Large Multimodal Models}. In \bibinfo{booktitle}{\emph{{ACL}}}. \bibinfo{publisher}{Association for Computational Linguistics}, \bibinfo{pages}{9663--9687}.
\newblock


\bibitem[Dai et~al\mbox{.}(2023)]%
        {InstructBLIP}
\bibfield{author}{\bibinfo{person}{Wenliang Dai}, \bibinfo{person}{Junnan Li}, \bibinfo{person}{Dongxu Li}, \bibinfo{person}{Anthony Meng~Huat Tiong}, \bibinfo{person}{Junqi Zhao}, \bibinfo{person}{Weisheng Wang}, \bibinfo{person}{Boyang Li}, \bibinfo{person}{Pascale Fung}, {and} \bibinfo{person}{Steven C.~H. Hoi}.} \bibinfo{year}{2023}\natexlab{}.
\newblock \showarticletitle{InstructBLIP: Towards General-purpose Vision-Language Models with Instruction Tuning}. In \bibinfo{booktitle}{\emph{NeurIPS}}.
\newblock


\bibitem[Desai et~al\mbox{.}(2022)]%
        {MORE}
\bibfield{author}{\bibinfo{person}{Poorav Desai}, \bibinfo{person}{Tanmoy Chakraborty}, {and} \bibinfo{person}{Md.~Shad Akhtar}.} \bibinfo{year}{2022}\natexlab{}.
\newblock \showarticletitle{Nice Perfume. How Long Did You Marinate in It? Multimodal Sarcasm Explanation}. In \bibinfo{booktitle}{\emph{AAAI}}. \bibinfo{publisher}{{AAAI} Press}, \bibinfo{pages}{10563--10571}.
\newblock


\bibitem[Ding et~al\mbox{.}(2022)]%
        {prompt-tuning}
\bibfield{author}{\bibinfo{person}{Daijun Ding}, \bibinfo{person}{Hu Huang}, \bibinfo{person}{Bowen Zhang}, \bibinfo{person}{Cheng Peng}, \bibinfo{person}{Yangyang Li}, \bibinfo{person}{Xianghua Fu}, {and} \bibinfo{person}{Liwen Jing}.} \bibinfo{year}{2022}\natexlab{}.
\newblock \showarticletitle{Multi-Modal Sarcasm Detection with Prompt-Tuning}.
\newblock
\href{https://doi.org/10.1109/ACAIT56212.2022.10137937}{doi:\nolinkurl{10.1109/ACAIT56212.2022.10137937}}


\bibitem[Fu et~al\mbox{.}(2023)]%
        {MME}
\bibfield{author}{\bibinfo{person}{Chaoyou Fu}, \bibinfo{person}{Peixian Chen}, \bibinfo{person}{Yunhang Shen}, \bibinfo{person}{Yulei Qin}, \bibinfo{person}{Mengdan Zhang}, \bibinfo{person}{Xu Lin}, \bibinfo{person}{Zhenyu Qiu}, \bibinfo{person}{Wei Lin}, \bibinfo{person}{Jinrui Yang}, \bibinfo{person}{Xiawu Zheng}, \bibinfo{person}{Ke Li}, \bibinfo{person}{Xing Sun}, {and} \bibinfo{person}{Rongrong Ji}.} \bibinfo{year}{2023}\natexlab{}.
\newblock \showarticletitle{{MME:} {A} Comprehensive Evaluation Benchmark for Multimodal Large Language Models}.
\newblock \bibinfo{journal}{\emph{CoRR}}  \bibinfo{volume}{abs/2306.13394} (\bibinfo{year}{2023}).
\newblock
\showeprint[arXiv]{2306.13394}


\bibitem[Girshick(2015)]%
        {FastRCNN}
\bibfield{author}{\bibinfo{person}{Ross~B. Girshick}.} \bibinfo{year}{2015}\natexlab{}.
\newblock \showarticletitle{Fast {R-CNN}}. In \bibinfo{booktitle}{\emph{{ICCV}}}. \bibinfo{publisher}{{IEEE} Computer Society}, \bibinfo{pages}{1440--1448}.
\newblock


\bibitem[Jing et~al\mbox{.}(2023a)]%
        {FaithScore}
\bibfield{author}{\bibinfo{person}{Liqiang Jing}, \bibinfo{person}{Ruosen Li}, \bibinfo{person}{Yunmo Chen}, \bibinfo{person}{Mengzhao Jia}, {and} \bibinfo{person}{Xinya Du}.} \bibinfo{year}{2023}\natexlab{a}.
\newblock \showarticletitle{{FAITHSCORE:} Evaluating Hallucinations in Large Vision-Language Models}.
\newblock \bibinfo{journal}{\emph{CoRR}}  \bibinfo{volume}{abs/2311.01477} (\bibinfo{year}{2023}).
\newblock
\showeprint[arXiv]{2311.01477}


\bibitem[Jing et~al\mbox{.}(2023b)]%
        {DBLP:journals/ijautcomp/JingLXYSS23}
\bibfield{author}{\bibinfo{person}{Liqiang Jing}, \bibinfo{person}{Yiren Li}, \bibinfo{person}{Junhao Xu}, \bibinfo{person}{Yongcan Yu}, \bibinfo{person}{Pei Shen}, {and} \bibinfo{person}{Xuemeng Song}.} \bibinfo{year}{2023}\natexlab{b}.
\newblock \showarticletitle{Vision Enhanced Generative Pre-trained Language Model for Multimodal Sentence Summarization}.
\newblock \bibinfo{journal}{\emph{Mach. Intell. Res.}} \bibinfo{volume}{20}, \bibinfo{number}{2} (\bibinfo{year}{2023}), \bibinfo{pages}{289--298}.
\newblock
\href{https://doi.org/10.1007/S11633-022-1372-X}{doi:\nolinkurl{10.1007/S11633-022-1372-X}}


\bibitem[Jing et~al\mbox{.}(2024)]%
        {sentiment4}
\bibfield{author}{\bibinfo{person}{Liqiang Jing}, \bibinfo{person}{Xuemeng Song}, \bibinfo{person}{Xuming Lin}, \bibinfo{person}{Zhongzhou Zhao}, \bibinfo{person}{Wei Zhou}, {and} \bibinfo{person}{Liqiang Nie}.} \bibinfo{year}{2024}\natexlab{}.
\newblock \showarticletitle{Stylized Data-to-text Generation: {A} Case Study in the E-Commerce Domain}.
\newblock \bibinfo{journal}{\emph{{ACM} Trans. Inf. Syst.}} \bibinfo{volume}{42}, \bibinfo{number}{1} (\bibinfo{year}{2024}), \bibinfo{pages}{25:1--25:24}.
\newblock
\href{https://doi.org/10.1145/3603374}{doi:\nolinkurl{10.1145/3603374}}


\bibitem[Jing et~al\mbox{.}(2023c)]%
        {TEAM}
\bibfield{author}{\bibinfo{person}{Liqiang Jing}, \bibinfo{person}{Xuemeng Song}, \bibinfo{person}{Kun Ouyang}, \bibinfo{person}{Mengzhao Jia}, {and} \bibinfo{person}{Liqiang Nie}.} \bibinfo{year}{2023}\natexlab{c}.
\newblock \showarticletitle{Multi-source Semantic Graph-based Multimodal Sarcasm Explanation Generation}. In \bibinfo{booktitle}{\emph{{ACL}}}. \bibinfo{publisher}{Association for Computational Linguistics}, \bibinfo{pages}{11349--11361}.
\newblock


\bibitem[Joshi et~al\mbox{.}(2016)]%
        {joshi-etal-2016-word}
\bibfield{author}{\bibinfo{person}{Aditya Joshi}, \bibinfo{person}{Vaibhav Tripathi}, \bibinfo{person}{Kevin Patel}, \bibinfo{person}{Pushpak Bhattacharyya}, {and} \bibinfo{person}{Mark Carman}.} \bibinfo{year}{2016}\natexlab{}.
\newblock \showarticletitle{Are Word Embedding-based Features Useful for Sarcasm Detection?}. In \bibinfo{booktitle}{\emph{EMNLP}}. \bibinfo{publisher}{Association for Computational Linguistics}, \bibinfo{pages}{1006--1011}.
\newblock


\bibitem[Li et~al\mbox{.}(2023)]%
        {Hallucination}
\bibfield{author}{\bibinfo{person}{Yifan Li}, \bibinfo{person}{Yifan Du}, \bibinfo{person}{Kun Zhou}, \bibinfo{person}{Jinpeng Wang}, \bibinfo{person}{Wayne~Xin Zhao}, {and} \bibinfo{person}{Ji{-}Rong Wen}.} \bibinfo{year}{2023}\natexlab{}.
\newblock \showarticletitle{Evaluating Object Hallucination in Large Vision-Language Models}. In \bibinfo{booktitle}{\emph{{EMNLP}}}. \bibinfo{publisher}{Association for Computational Linguistics}, \bibinfo{pages}{292--305}.
\newblock


\bibitem[Lin(2004)]%
        {ROUGE}
\bibfield{author}{\bibinfo{person}{Chin-Yew Lin}.} \bibinfo{year}{2004}\natexlab{}.
\newblock \showarticletitle{{ROUGE}: A Package for Automatic Evaluation of Summaries}. In \bibinfo{booktitle}{\emph{Text Summarization Branches Out}}. \bibinfo{publisher}{Association for Computational Linguistics}, \bibinfo{pages}{74--81}.
\newblock


\bibitem[Lin et~al\mbox{.}(2023)]%
        {DBLP:conf/sigir/LinJSLSN23}
\bibfield{author}{\bibinfo{person}{Dengtian Lin}, \bibinfo{person}{Liqiang Jing}, \bibinfo{person}{Xuemeng Song}, \bibinfo{person}{Meng Liu}, \bibinfo{person}{Teng Sun}, {and} \bibinfo{person}{Liqiang Nie}.} \bibinfo{year}{2023}\natexlab{}.
\newblock \showarticletitle{Adapting Generative Pretrained Language Model for Open-domain Multimodal Sentence Summarization}. In \bibinfo{booktitle}{\emph{Proceedings of the 46th International {ACM} {SIGIR} Conference on Research and Development in Information Retrieval, {SIGIR} 2023, Taipei, Taiwan, July 23-27, 2023}}, \bibfield{editor}{\bibinfo{person}{Hsin{-}Hsi Chen}, \bibinfo{person}{Wei{-}Jou~(Edward) Duh}, \bibinfo{person}{Hen{-}Hsen Huang}, \bibinfo{person}{Makoto~P. Kato}, \bibinfo{person}{Josiane Mothe}, {and} \bibinfo{person}{Barbara Poblete}} (Eds.). \bibinfo{publisher}{{ACM}}, \bibinfo{pages}{195--204}.
\newblock
\href{https://doi.org/10.1145/3539618.3591633}{doi:\nolinkurl{10.1145/3539618.3591633}}


\bibitem[Lin et~al\mbox{.}(2024)]%
        {Goat-Bench}
\bibfield{author}{\bibinfo{person}{Hongzhan Lin}, \bibinfo{person}{Ziyang Luo}, \bibinfo{person}{Bo Wang}, \bibinfo{person}{Ruichao Yang}, {and} \bibinfo{person}{Jing Ma}.} \bibinfo{year}{2024}\natexlab{}.
\newblock \showarticletitle{GOAT-Bench: Safety Insights to Large Multimodal Models through Meme-Based Social Abuse}.
\newblock \bibinfo{journal}{\emph{CoRR}}  \bibinfo{volume}{abs/2401.01523} (\bibinfo{year}{2024}).
\newblock


\bibitem[Liu et~al\mbox{.}(2023)]%
        {LLaVA}
\bibfield{author}{\bibinfo{person}{Haotian Liu}, \bibinfo{person}{Chunyuan Li}, \bibinfo{person}{Qingyang Wu}, {and} \bibinfo{person}{Yong~Jae Lee}.} \bibinfo{year}{2023}\natexlab{}.
\newblock \showarticletitle{Visual Instruction Tuning}. In \bibinfo{booktitle}{\emph{NeurIPS}}.
\newblock


\bibitem[Liu et~al\mbox{.}(2025)]%
        {DBLP:journals/corr/abs-2504-21226}
\bibfield{author}{\bibinfo{person}{Jiaqi Liu}, \bibinfo{person}{Ran Tong}, \bibinfo{person}{Aowei Shen}, \bibinfo{person}{Shuzheng Li}, \bibinfo{person}{Changlin Yang}, {and} \bibinfo{person}{Lisha Xu}.} \bibinfo{year}{2025}\natexlab{}.
\newblock \showarticletitle{MemeBLIP2: {A} novel lightweight multimodal system to detect harmful memes}.
\newblock \bibinfo{journal}{\emph{CoRR}}  \bibinfo{volume}{abs/2504.21226} (\bibinfo{year}{2025}).
\newblock
\href{https://doi.org/10.48550/ARXIV.2504.21226}{doi:\nolinkurl{10.48550/ARXIV.2504.21226}}
\showeprint[arXiv]{2504.21226}


\bibitem[Liu et~al\mbox{.}(2024)]%
        {FakeNews}
\bibfield{author}{\bibinfo{person}{Xuannan Liu}, \bibinfo{person}{Peipei Li}, \bibinfo{person}{Huaibo Huang}, \bibinfo{person}{Zekun Li}, \bibinfo{person}{Xing Cui}, \bibinfo{person}{Jiahao Liang}, \bibinfo{person}{Lixiong Qin}, \bibinfo{person}{Weihong Deng}, {and} \bibinfo{person}{Zhaofeng He}.} \bibinfo{year}{2024}\natexlab{}.
\newblock \showarticletitle{FakeNewsGPT4: Advancing Multimodal Fake News Detection through Knowledge-Augmented LVLMs}.
\newblock \bibinfo{journal}{\emph{CoRR}}  \bibinfo{volume}{abs/2403.01988} (\bibinfo{year}{2024}).
\newblock
\showeprint[arXiv]{2403.01988}


\bibitem[Lu et~al\mbox{.}(2023)]%
        {lu2023evaluation}
\bibfield{author}{\bibinfo{person}{Jiaying Lu}, \bibinfo{person}{Jinmeng Rao}, \bibinfo{person}{Kezhen Chen}, \bibinfo{person}{Xiaoyuan Guo}, \bibinfo{person}{Yawen Zhang}, \bibinfo{person}{Baochen Sun}, \bibinfo{person}{Carl Yang}, {and} \bibinfo{person}{Jie Yang}.} \bibinfo{year}{2023}\natexlab{}.
\newblock \bibinfo{title}{Evaluation and Enhancement of Semantic Grounding in Large Vision-Language Models}.
\newblock
\showeprint[arxiv]{2309.04041}~[cs.CV]


\bibitem[OpenAI(2023)]%
        {GPT4}
\bibfield{author}{\bibinfo{person}{OpenAI}.} \bibinfo{year}{2023}\natexlab{}.
\newblock \showarticletitle{{GPT-4} Technical Report}.
\newblock \bibinfo{journal}{\emph{CoRR}}  \bibinfo{volume}{abs/2303.08774} (\bibinfo{year}{2023}).
\newblock
\showeprint[arXiv]{2303.08774}


\bibitem[Ouyang et~al\mbox{.}(2024)]%
        {DBLP:journals/corr/abs-2402-03658}
\bibfield{author}{\bibinfo{person}{Kun Ouyang}, \bibinfo{person}{Liqiang Jing}, \bibinfo{person}{Xuemeng Song}, \bibinfo{person}{Meng Liu}, \bibinfo{person}{Yupeng Hu}, {and} \bibinfo{person}{Liqiang Nie}.} \bibinfo{year}{2024}\natexlab{}.
\newblock \showarticletitle{Sentiment-enhanced Graph-based Sarcasm Explanation in Dialogue}.
\newblock \bibinfo{journal}{\emph{CoRR}}  \bibinfo{volume}{abs/2402.03658} (\bibinfo{year}{2024}).
\newblock
\href{https://doi.org/10.48550/ARXIV.2402.03658}{doi:\nolinkurl{10.48550/ARXIV.2402.03658}}
\showeprint[arXiv]{2402.03658}


\bibitem[Papineni et~al\mbox{.}(2002)]%
        {BLEU}
\bibfield{author}{\bibinfo{person}{Kishore Papineni}, \bibinfo{person}{Salim Roukos}, \bibinfo{person}{Todd Ward}, {and} \bibinfo{person}{Wei{-}Jing Zhu}.} \bibinfo{year}{2002}\natexlab{}.
\newblock \showarticletitle{Bleu: a Method for Automatic Evaluation of Machine Translation}. In \bibinfo{booktitle}{\emph{ACL}}. \bibinfo{publisher}{{ACL}}, \bibinfo{pages}{311--318}.
\newblock


\bibitem[Qiao et~al\mbox{.}(2023)]%
        {Incongruity}
\bibfield{author}{\bibinfo{person}{Yang Qiao}, \bibinfo{person}{Liqiang Jing}, \bibinfo{person}{Xuemeng Song}, \bibinfo{person}{Xiaolin Chen}, \bibinfo{person}{Lei Zhu}, {and} \bibinfo{person}{Liqiang Nie}.} \bibinfo{year}{2023}\natexlab{}.
\newblock \showarticletitle{Mutual-Enhanced Incongruity Learning Network for Multi-Modal Sarcasm Detection}. In \bibinfo{booktitle}{\emph{AAAI}}. \bibinfo{publisher}{{AAAI} Press}, \bibinfo{pages}{9507--9515}.
\newblock


\bibitem[Schifanella et~al\mbox{.}(2016)]%
        {MSD}
\bibfield{author}{\bibinfo{person}{Rossano Schifanella}, \bibinfo{person}{Paloma de Juan}, \bibinfo{person}{Joel~R. Tetreault}, {and} \bibinfo{person}{Liangliang Cao}.} \bibinfo{year}{2016}\natexlab{}.
\newblock \showarticletitle{Detecting Sarcasm in Multimodal Social Platforms}. In \bibinfo{booktitle}{\emph{{MM} 2016}}. \bibinfo{publisher}{{ACM}}, \bibinfo{pages}{1136--1145}.
\newblock


\bibitem[Shukor et~al\mbox{.}(2024)]%
        {EvalICL}
\bibfield{author}{\bibinfo{person}{Mustafa Shukor}, \bibinfo{person}{Alexandre Ram{\'{e}}}, \bibinfo{person}{Corentin Dancette}, {and} \bibinfo{person}{Matthieu Cord}.} \bibinfo{year}{2024}\natexlab{}.
\newblock \showarticletitle{Beyond task performance: evaluating and reducing the flaws of large multimodal models with in-context-learning}. In \bibinfo{booktitle}{\emph{{ICLR} 2024}}. \bibinfo{publisher}{OpenReview.net}.
\newblock


\bibitem[Song et~al\mbox{.}(2022)]%
        {DBLP:conf/sigir/SongJLZCN22}
\bibfield{author}{\bibinfo{person}{Xuemeng Song}, \bibinfo{person}{Liqiang Jing}, \bibinfo{person}{Dengtian Lin}, \bibinfo{person}{Zhongzhou Zhao}, \bibinfo{person}{Haiqing Chen}, {and} \bibinfo{person}{Liqiang Nie}.} \bibinfo{year}{2022}\natexlab{}.
\newblock \showarticletitle{{V2P:} Vision-to-Prompt based Multi-Modal Product Summary Generation}. In \bibinfo{booktitle}{\emph{{SIGIR} '22: The 45th International {ACM} {SIGIR} Conference on Research and Development in Information Retrieval, Madrid, Spain, July 11 - 15, 2022}}, \bibfield{editor}{\bibinfo{person}{Enrique Amig{\'{o}}}, \bibinfo{person}{Pablo Castells}, \bibinfo{person}{Julio Gonzalo}, \bibinfo{person}{Ben Carterette}, \bibinfo{person}{J.~Shane Culpepper}, {and} \bibinfo{person}{Gabriella Kazai}} (Eds.). \bibinfo{publisher}{{ACM}}, \bibinfo{pages}{992--1001}.
\newblock
\href{https://doi.org/10.1145/3477495.3532076}{doi:\nolinkurl{10.1145/3477495.3532076}}


\bibitem[Speer et~al\mbox{.}(2017)]%
        {ConceptNet}
\bibfield{author}{\bibinfo{person}{Robyn Speer}, \bibinfo{person}{Joshua Chin}, {and} \bibinfo{person}{Catherine Havasi}.} \bibinfo{year}{2017}\natexlab{}.
\newblock \showarticletitle{ConceptNet 5.5: An Open Multilingual Graph of General Knowledge}. In \bibinfo{booktitle}{\emph{{AAAI}}}. \bibinfo{publisher}{{AAAI} Press}, \bibinfo{pages}{4444--4451}.
\newblock


\bibitem[Sun et~al\mbox{.}(2023a)]%
        {sentiment1}
\bibfield{author}{\bibinfo{person}{Teng Sun}, \bibinfo{person}{Liqiang Jing}, \bibinfo{person}{Yinwei Wei}, \bibinfo{person}{Xuemeng Song}, \bibinfo{person}{Zhiyong Cheng}, {and} \bibinfo{person}{Liqiang Nie}.} \bibinfo{year}{2023}\natexlab{a}.
\newblock \showarticletitle{Dual Consistency-Enhanced Semi-Supervised Sentiment Analysis Towards {COVID-19} Tweets}.
\newblock \bibinfo{journal}{\emph{{IEEE} Trans. Knowl. Data Eng.}} \bibinfo{volume}{35}, \bibinfo{number}{12} (\bibinfo{year}{2023}), \bibinfo{pages}{12605--12617}.
\newblock
\href{https://doi.org/10.1109/TKDE.2023.3270940}{doi:\nolinkurl{10.1109/TKDE.2023.3270940}}


\bibitem[Sun et~al\mbox{.}(2023b)]%
        {sentiment3}
\bibfield{author}{\bibinfo{person}{Teng Sun}, \bibinfo{person}{Juntong Ni}, \bibinfo{person}{Wenjie Wang}, \bibinfo{person}{Liqiang Jing}, \bibinfo{person}{Yinwei Wei}, {and} \bibinfo{person}{Liqiang Nie}.} \bibinfo{year}{2023}\natexlab{b}.
\newblock \showarticletitle{General Debiasing for Multimodal Sentiment Analysis}. In \bibinfo{booktitle}{\emph{Proceedings of the 31st {ACM} International Conference on Multimedia, {MM} 2023, Ottawa, ON, Canada, 29 October 2023- 3 November 2023}}, \bibfield{editor}{\bibinfo{person}{Abdulmotaleb El{-}Saddik}, \bibinfo{person}{Tao Mei}, \bibinfo{person}{Rita Cucchiara}, \bibinfo{person}{Marco Bertini}, \bibinfo{person}{Diana Patricia~Tobon Vallejo}, \bibinfo{person}{Pradeep~K. Atrey}, {and} \bibinfo{person}{M.~Shamim Hossain}} (Eds.). \bibinfo{publisher}{{ACM}}, \bibinfo{pages}{5861--5869}.
\newblock
\href{https://doi.org/10.1145/3581783.3612051}{doi:\nolinkurl{10.1145/3581783.3612051}}


\bibitem[Sun et~al\mbox{.}(2022)]%
        {sentiment2}
\bibfield{author}{\bibinfo{person}{Teng Sun}, \bibinfo{person}{Wenjie Wang}, \bibinfo{person}{Liqiang Jing}, \bibinfo{person}{Yiran Cui}, \bibinfo{person}{Xuemeng Song}, {and} \bibinfo{person}{Liqiang Nie}.} \bibinfo{year}{2022}\natexlab{}.
\newblock \showarticletitle{Counterfactual Reasoning for Out-of-distribution Multimodal Sentiment Analysis}. In \bibinfo{booktitle}{\emph{{MM} '22: The 30th {ACM} International Conference on Multimedia, Lisboa, Portugal, October 10 - 14, 2022}}, \bibfield{editor}{\bibinfo{person}{Jo{\~{a}}o Magalh{\~{a}}es}, \bibinfo{person}{Alberto~Del Bimbo}, \bibinfo{person}{Shin'ichi Satoh}, \bibinfo{person}{Nicu Sebe}, \bibinfo{person}{Xavier Alameda{-}Pineda}, \bibinfo{person}{Qin Jin}, \bibinfo{person}{Vincent Oria}, {and} \bibinfo{person}{Laura Toni}} (Eds.). \bibinfo{publisher}{{ACM}}, \bibinfo{pages}{15--23}.
\newblock
\href{https://doi.org/10.1145/3503161.3548211}{doi:\nolinkurl{10.1145/3503161.3548211}}


\bibitem[Tang et~al\mbox{.}(2024)]%
        {DBLP:conf/naacl/TangLY024}
\bibfield{author}{\bibinfo{person}{Binghao Tang}, \bibinfo{person}{Boda Lin}, \bibinfo{person}{Haolong Yan}, {and} \bibinfo{person}{Si Li}.} \bibinfo{year}{2024}\natexlab{}.
\newblock \showarticletitle{Leveraging Generative Large Language Models with Visual Instruction and Demonstration Retrieval for Multimodal Sarcasm Detection}. In \bibinfo{booktitle}{\emph{{NAACL}}}. \bibinfo{publisher}{Association for Computational Linguistics}, \bibinfo{pages}{1732--1742}.
\newblock


\bibitem[Tong et~al\mbox{.}(2025)]%
        {tong2025rainbow}
\bibfield{author}{\bibinfo{person}{Ran Tong}, \bibinfo{person}{Songtao Wei}, \bibinfo{person}{Jiaqi Liu}, {and} \bibinfo{person}{Lanruo Wang}.} \bibinfo{year}{2025}\natexlab{}.
\newblock \showarticletitle{Rainbow Noise: Stress-Testing Multimodal Harmful-Meme Detectors on LGBTQ Content}.
\newblock \bibinfo{journal}{\emph{arXiv preprint arXiv:2507.19551}} (\bibinfo{year}{2025}).
\newblock


\bibitem[Touvron et~al\mbox{.}(2023)]%
        {LLaMA}
\bibfield{author}{\bibinfo{person}{Hugo Touvron}, \bibinfo{person}{Thibaut Lavril}, \bibinfo{person}{Gautier Izacard}, \bibinfo{person}{Xavier Martinet}, \bibinfo{person}{Marie{-}Anne Lachaux}, \bibinfo{person}{Timoth{\'{e}}e Lacroix}, \bibinfo{person}{Baptiste Rozi{\`{e}}re}, \bibinfo{person}{Naman Goyal}, \bibinfo{person}{Eric Hambro}, \bibinfo{person}{Faisal Azhar}, \bibinfo{person}{Aur{\'{e}}lien Rodriguez}, \bibinfo{person}{Armand Joulin}, \bibinfo{person}{Edouard Grave}, {and} \bibinfo{person}{Guillaume Lample}.} \bibinfo{year}{2023}\natexlab{}.
\newblock \showarticletitle{LLaMA: Open and Efficient Foundation Language Models}.
\newblock \bibinfo{journal}{\emph{CoRR}}  \bibinfo{volume}{abs/2302.13971} (\bibinfo{year}{2023}).
\newblock
\showeprint[arXiv]{2302.13971}


\bibitem[Wang et~al\mbox{.}(2024)]%
        {Wisdom}
\bibfield{author}{\bibinfo{person}{Wenbin Wang}, \bibinfo{person}{Liang Ding}, \bibinfo{person}{Li Shen}, \bibinfo{person}{Yong Luo}, \bibinfo{person}{Han Hu}, {and} \bibinfo{person}{Dacheng Tao}.} \bibinfo{year}{2024}\natexlab{}.
\newblock \showarticletitle{WisdoM: Improving Multimodal Sentiment Analysis by Fusing Contextual World Knowledge}.
\newblock \bibinfo{journal}{\emph{CoRR}}  \bibinfo{volume}{abs/2401.06659} (\bibinfo{year}{2024}).
\newblock
\showeprint[arXiv]{2401.06659}


\bibitem[Wen et~al\mbox{.}(2023)]%
        {Wen_2023_CVPR}
\bibfield{author}{\bibinfo{person}{Changsong Wen}, \bibinfo{person}{Guoli Jia}, {and} \bibinfo{person}{Jufeng Yang}.} \bibinfo{year}{2023}\natexlab{}.
\newblock \showarticletitle{DIP: Dual Incongruity Perceiving Network for Sarcasm Detection}. In \bibinfo{booktitle}{\emph{CVPR}}. \bibinfo{pages}{2540--2550}.
\newblock


\bibitem[Wu et~al\mbox{.}(2021)]%
        {ModelIncongruity}
\bibfield{author}{\bibinfo{person}{Yang Wu}, \bibinfo{person}{Yanyan Zhao}, \bibinfo{person}{Xin Lu}, \bibinfo{person}{Bing Qin}, \bibinfo{person}{Yin Wu}, \bibinfo{person}{Jian Sheng}, {and} \bibinfo{person}{Jinlong Li}.} \bibinfo{year}{2021}\natexlab{}.
\newblock \showarticletitle{Modeling Incongruity between Modalities for Multimodal Sarcasm Detection}.
\newblock \bibinfo{journal}{\emph{{IEEE} Multim.}} \bibinfo{volume}{28}, \bibinfo{number}{2} (\bibinfo{year}{2021}), \bibinfo{pages}{86--95}.
\newblock


\bibitem[Xu et~al\mbox{.}(2023)]%
        {LVLM-eHub}
\bibfield{author}{\bibinfo{person}{Peng Xu}, \bibinfo{person}{Wenqi Shao}, \bibinfo{person}{Kaipeng Zhang}, \bibinfo{person}{Peng Gao}, \bibinfo{person}{Shuo Liu}, \bibinfo{person}{Meng Lei}, \bibinfo{person}{Fanqing Meng}, \bibinfo{person}{Siyuan Huang}, \bibinfo{person}{Yu Qiao}, {and} \bibinfo{person}{Ping Luo}.} \bibinfo{year}{2023}\natexlab{}.
\newblock \showarticletitle{LVLM-eHub: {A} Comprehensive Evaluation Benchmark for Large Vision-Language Models}.
\newblock \bibinfo{journal}{\emph{CoRR}}  \bibinfo{volume}{abs/2306.09265} (\bibinfo{year}{2023}).
\newblock
\showeprint[arXiv]{2306.09265}


\bibitem[Xuan et~al\mbox{.}(2024)]%
        {LEMMA}
\bibfield{author}{\bibinfo{person}{Keyang Xuan}, \bibinfo{person}{Li Yi}, \bibinfo{person}{Fan Yang}, \bibinfo{person}{Ruochen Wu}, \bibinfo{person}{Yi~R. Fung}, {and} \bibinfo{person}{Heng Ji}.} \bibinfo{year}{2024}\natexlab{}.
\newblock \showarticletitle{{LEMMA:} Towards LVLM-Enhanced Multimodal Misinformation Detection with External Knowledge Augmentation}.
\newblock \bibinfo{journal}{\emph{CoRR}}  \bibinfo{volume}{abs/2402.11943} (\bibinfo{year}{2024}).
\newblock
\showeprint[arXiv]{2402.11943}


\bibitem[Yang et~al\mbox{.}(2024)]%
        {MM-Instruct}
\bibfield{author}{\bibinfo{person}{Xiaocui Yang}, \bibinfo{person}{Wenfang Wu}, \bibinfo{person}{Shi Feng}, \bibinfo{person}{Ming Wang}, \bibinfo{person}{Daling Wang}, \bibinfo{person}{Yang Li}, \bibinfo{person}{Qi Sun}, \bibinfo{person}{Yifei Zhang}, \bibinfo{person}{Xiaoming Fu}, {and} \bibinfo{person}{Soujanya Poria}.} \bibinfo{year}{2024}\natexlab{}.
\newblock \showarticletitle{MM-InstructEval: Zero-Shot Evaluation of (Multimodal) Large Language Models on Multimodal Reasoning Tasks}.
\newblock \bibinfo{journal}{\emph{CoRR}}  \bibinfo{volume}{abs/2405.07229} (\bibinfo{year}{2024}).
\newblock
\showeprint[arXiv]{2405.07229}


\bibitem[Yue et~al\mbox{.}(2023)]%
        {KnowleNet}
\bibfield{author}{\bibinfo{person}{Tan Yue}, \bibinfo{person}{Rui Mao}, \bibinfo{person}{Heng Wang}, \bibinfo{person}{Zonghai Hu}, {and} \bibinfo{person}{Erik Cambria}.} \bibinfo{year}{2023}\natexlab{}.
\newblock \showarticletitle{KnowleNet: Knowledge fusion network for multimodal sarcasm detection}.
\newblock \bibinfo{journal}{\emph{Inf. Fusion}}  \bibinfo{volume}{100} (\bibinfo{year}{2023}), \bibinfo{pages}{101921}.
\newblock


\bibitem[Zhang et~al\mbox{.}(2025a)]%
        {DBLP:conf/aaai/ZhangJG25}
\bibfield{author}{\bibinfo{person}{Yue Zhang}, \bibinfo{person}{Liqiang Jing}, {and} \bibinfo{person}{Vibhav Gogate}.} \bibinfo{year}{2025}\natexlab{a}.
\newblock \showarticletitle{Defeasible Visual Entailment: Benchmark, Evaluator, and Reward-Driven Optimization}. In \bibinfo{booktitle}{\emph{AAAI-25, Sponsored by the Association for the Advancement of Artificial Intelligence, February 25 - March 4, 2025, Philadelphia, PA, {USA}}}, \bibfield{editor}{\bibinfo{person}{Toby Walsh}, \bibinfo{person}{Julie Shah}, {and} \bibinfo{person}{Zico Kolter}} (Eds.). \bibinfo{publisher}{{AAAI} Press}, \bibinfo{pages}{25976--25984}.
\newblock
\href{https://doi.org/10.1609/AAAI.V39I24.34792}{doi:\nolinkurl{10.1609/AAAI.V39I24.34792}}


\bibitem[Zhang et~al\mbox{.}(2025b)]%
        {DBLP:journals/corr/abs-2506-22385}
\bibfield{author}{\bibinfo{person}{Yue Zhang}, \bibinfo{person}{Jilei Sun}, \bibinfo{person}{Yunhui Guo}, {and} \bibinfo{person}{Vibhav Gogate}.} \bibinfo{year}{2025}\natexlab{b}.
\newblock \showarticletitle{Can Video Large Multimodal Models Think Like Doubters-or Double-Down: {A} Study on Defeasible Video Entailment}.
\newblock \bibinfo{journal}{\emph{CoRR}}  \bibinfo{volume}{abs/2506.22385} (\bibinfo{year}{2025}).
\newblock
\href{https://doi.org/10.48550/ARXIV.2506.22385}{doi:\nolinkurl{10.48550/ARXIV.2506.22385}}
\showeprint[arXiv]{2506.22385}


\bibitem[Zhang et~al\mbox{.}(2024b)]%
        {zhang2024fine}
\bibfield{author}{\bibinfo{person}{Yue Zhang}, \bibinfo{person}{Jingxuan Zuo}, {and} \bibinfo{person}{Liqiang Jing}.} \bibinfo{year}{2024}\natexlab{b}.
\newblock \showarticletitle{Fine-grained and explainable factuality evaluation for multimodal summarization}.
\newblock \bibinfo{journal}{\emph{arXiv preprint arXiv:2402.11414}} (\bibinfo{year}{2024}).
\newblock


\bibitem[Zhang et~al\mbox{.}(2024a)]%
        {MM-CoT}
\bibfield{author}{\bibinfo{person}{Zhuosheng Zhang}, \bibinfo{person}{Aston Zhang}, \bibinfo{person}{Mu Li}, \bibinfo{person}{Hai Zhao}, \bibinfo{person}{George Karypis}, {and} \bibinfo{person}{Alex Smola}.} \bibinfo{year}{2024}\natexlab{a}.
\newblock \showarticletitle{Multimodal Chain-of-Thought Reasoning in Language Models}.
\newblock \bibinfo{journal}{\emph{Trans. Mach. Learn. Res.}}  \bibinfo{volume}{2024} (\bibinfo{year}{2024}).
\newblock


\bibitem[Zheng et~al\mbox{.}(2023)]%
        {DDCOT}
\bibfield{author}{\bibinfo{person}{Ge Zheng}, \bibinfo{person}{Bin Yang}, \bibinfo{person}{Jiajin Tang}, \bibinfo{person}{Hong{-}Yu Zhou}, {and} \bibinfo{person}{Sibei Yang}.} \bibinfo{year}{2023}\natexlab{}.
\newblock \showarticletitle{DDCoT: Duty-Distinct Chain-of-Thought Prompting for Multimodal Reasoning in Language Models}. In \bibinfo{booktitle}{\emph{NeurIPS 2023}}.
\newblock


\bibitem[Zhou et~al\mbox{.}(2024a)]%
        {AnalyzeHallucination}
\bibfield{author}{\bibinfo{person}{Yiyang Zhou}, \bibinfo{person}{Chenhang Cui}, \bibinfo{person}{Jaehong Yoon}, \bibinfo{person}{Linjun Zhang}, \bibinfo{person}{Zhun Deng}, \bibinfo{person}{Chelsea Finn}, \bibinfo{person}{Mohit Bansal}, {and} \bibinfo{person}{Huaxiu Yao}.} \bibinfo{year}{2024}\natexlab{a}.
\newblock \showarticletitle{Analyzing and Mitigating Object Hallucination in Large Vision-Language Models}. In \bibinfo{booktitle}{\emph{{ICLR} 2024}}. \bibinfo{publisher}{OpenReview.net}.
\newblock


\bibitem[Zhou et~al\mbox{.}(2024b)]%
        {VICL}
\bibfield{author}{\bibinfo{person}{Yucheng Zhou}, \bibinfo{person}{Xiang Li}, \bibinfo{person}{Qianning Wang}, {and} \bibinfo{person}{Jianbing Shen}.} \bibinfo{year}{2024}\natexlab{b}.
\newblock \showarticletitle{Visual In-Context Learning for Large Vision-Language Models}. In \bibinfo{booktitle}{\emph{Findings of the Association for Computational Linguistics, {ACL} 2024}}. \bibinfo{publisher}{Association for Computational Linguistics}, \bibinfo{pages}{15890--15902}.
\newblock


\bibitem[Zhu et~al\mbox{.}(2024)]%
        {MiniGPT4}
\bibfield{author}{\bibinfo{person}{Deyao Zhu}, \bibinfo{person}{Jun Chen}, \bibinfo{person}{Xiaoqian Shen}, \bibinfo{person}{Xiang Li}, {and} \bibinfo{person}{Mohamed Elhoseiny}.} \bibinfo{year}{2024}\natexlab{}.
\newblock \showarticletitle{MiniGPT-4: Enhancing Vision-Language Understanding with Advanced Large Language Models}. In \bibinfo{booktitle}{\emph{{ICLR}}}. \bibinfo{publisher}{OpenReview.net}.
\newblock


\end{thebibliography}
